\documentclass[letterpaper]{article} 
\usepackage{aaai24}  
\usepackage{times}  
\usepackage{helvet}  
\usepackage{courier}  
\usepackage[hyphens]{url}  
\usepackage{graphicx} 
\usepackage{natbib}  
\usepackage{caption} 
\usepackage{algorithm}
\usepackage{algorithmic}

\usepackage{newfloat}
\usepackage{listings}
\DeclareCaptionStyle{ruled}{labelfont=normalfont,labelsep=colon,strut=off} 
\floatstyle{ruled}
\newfloat{listing}{tb}{lst}{}
\floatname{listing}{Listing}
\usepackage{amsmath}
\usepackage{amssymb}
\usepackage{multirow}
\usepackage{arydshln}

\title{HyperLoRA for PDEs}
\author {
    Ritam Majumdar \textsuperscript{\rm 1},
    Vishal Jadhav \textsuperscript{\rm 1},
    Anirudh Deodhar \textsuperscript{\rm 1},
    Shirish Karande \textsuperscript{\rm 1},
    Lovekesh Vig \textsuperscript{\rm 1},
    Venkataramana Runkana \textsuperscript{\rm 1}
}
\affiliations {
    \textsuperscript{\rm 1}Tata Consultancy Services Research\\
    \{ritam.majumdar, 
    vi.suja,
    anirudh.deodhar,
    shirish.karande,
    lovekesh.vig,
    venkat.runkana\}@tcs.com
}

\usepackage{bibentry}

\begin{document}

\maketitle

\begin{abstract}
Physics-informed neural networks (PINNs) have been widely used to develop neural surrogates for solutions of Partial Differential Equations. A drawback of PINNs is that they have to be retrained with every change in initial-boundary conditions and PDE coefficients. The Hypernetwork, a model-based meta learning technique, takes in a parameterized task embedding as input and predicts the weights of PINN as output. Predicting weights of a neural network however, is a high-dimensional regression problem, and hypernetworks perform sub-optimally while predicting parameters for large base networks. To circumvent this issue, we use a low ranked adaptation (LoRA) formulation to decompose every layer of the base network into low-ranked tensors and use hypernetworks to predict the low-ranked tensors. Despite the reduced dimensionality of the resulting weight-regression problem, LoRA-based Hypernetworks violate the underlying physics of the given task. We demonstrate that the generalization capabilities of LoRA-based hypernetworks drastically improve when trained with an additional physics-informed loss component (HyperPINN) to satisfy the governing differential equations. We observe that LoRA-based HyperPINN training allows us to learn fast solutions for parameterized PDEs like Burger's equation and Navier Stokes: Kovasznay flow, while having an 8x reduction in prediction parameters on average without  compromising on accuracy when compared to all other baselines.
\end{abstract}

\section{Introduction}
Partial Differential Equations have been used to model several physical phenomenon, with applications cutting across Industries, including but not limited to Chemical, Aerospace, Automotive, Power, Pharmaceuticals and Metallurgy. However, most systems of PDEs are not amenable to analytical solutions and require numerical approximations which are very time consuming. Additionally, even the slightest change in the initial or boundary value conditions, or the coefficient values of a PDE, may require running fresh computational simulations. These limitations have spurred significant interest in applying neural networks for accelerating numerical methods \cite{RAISSI2019686, kochkov2021machine,greenfeld2019learning}.

A technique for obtaining a neural network based fast solver is to approximate the operator using data driven methods. Early approaches such as \cite{10.1145/2939672.2939738} assumed a finite dimensional fixed resolution mesh for approximating the solution. However recent advances on Neural Operators \cite{lu2019deeponet, li2020fourier} allow for mesh-free learning by utilizing a single set of network parameters for different discretizations. Meanwhile, there is a recent trend to represent data with implicit neural representatons (INRs) and these have been used to represent  images \cite{ha2016generating}, and 3D shapes \cite{Mescheder_2019_CVPR}. There is interest in performing deeplearning tasks on a dataset formed by such INRs \cite{dupont2022data}. Such an approach can be used for obtaining fast solvers as well, where a single solution is represented by a neural network and yet another neural network is trained by treating such INRs as data. Such Hyper-networks learn to directly predict weights of an INR for unseen  conditions and PDEs. In case of PDEs the dataset of INRs can also be obtained by employing a solver which directly obtains neural surrogates for the solution of a PDE \cite{majumdar2022realtime}.

Hypernetwork \cite{ha2016hypernetworks} is a form model based meta learning network that takes task descriptions as input and predicts the weights of a neural network as output. Predicting weights of a neural network however, is a high-dimensional regression problem, and hypernetworks perform sub-optimally while predicting parameters for large base networks. However, one can represent the adaptation of large network with parameter efficient tuning \cite{hu2021lora,li2021prefixtuning}. Thus, a hypernetwork can be trained using just adaptation parameters rather than the entire network. Almost concurrent to our work such ideas have been used for Image Generation \cite{ruiz2023hyperdreambooth} and Instruction Tuning of LLMs \cite{ivison2023hint}.  

In this work we show the efficacy of utilizing Hypernetworks to predict low rank adaptation weights of neural surrogate for solving PDEs. We observe that hypernetworks trained purely by using INRs as data do not provide the best performance and significant performance gains can be obtained if one has access to input-output data for the neural surrogates. Such data may be available from sensors or from employing a PINN solver for each task. Further, we observed the best performance when a physics informed loss is included during hypernetwork training. 

The remainder of the paper is organized as follows. First, we mathematically define the LoRA-based equations in the methodology section of the paper. We follow it up by Dataset information, training and hyperparameter details. We discuss the results and observations in the next section and summarize our findings in the conclusions.  
\section{Methodology}

\begin{figure*}
    \centering
    \includegraphics[height=6cm, width=\textwidth]{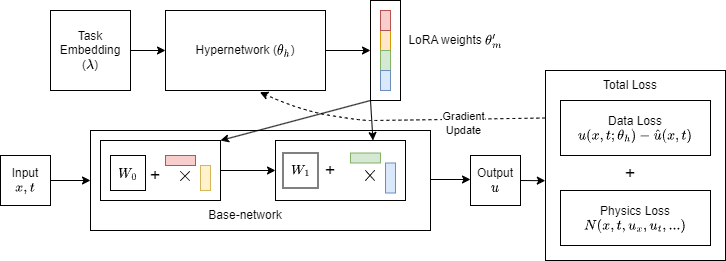}
    \caption{LoRA-based HyperPINN. The hypernetwork parameterized by $\theta_h$, takes in task embedding $\lambda$ as input, and outputs Low-ranked weights $\theta^\prime_m$. They are passed into the base-network (We show 2-layers as reference) with pre-trained weights $W_0,W_1$. The base network takes in variables of PDE $x,t$ as input and outputs $u$, which is trained in a physics-informed manner to update $\theta_h$.}
    \label{fig:HyperPINNLoRA}
\end{figure*}

\subsection{Physics-informed Neural Networks}
We consider Partial Differential Equations of the form 
\begin{equation}
N[x,t,u(x,t)]=0, t\in[0,T], x\in\Omega 
\end{equation}
Here, $N$ is a non-linear differential operator consisting of space and time derivatives of spatial variable $x$ and temporal variable $t$. The neural network is trained using a loss function that consists of: 1) A supervised initial condition loss $L_{IC}$ and boundary condition loss $L_{BC}$ component and 2) An unsupervised physics-informed loss component. 

\begin{align}
\label{PINN}
L(\theta) &= L_{IC}(\theta) + L_{BC}(\theta) + L_{Physics}(\theta) \nonumber\\
L_{IC}(\theta) &= \sum_{x_i \in IC} (u(x_i,0)-f(x_i))^2  \nonumber\\
L_{BC}(\theta) &= \sum_{x_{bc},t_{bc} \in BC} (u(x_{bc},t_{bc})-h(x_{bc},t_{bc}))^2\nonumber\\
L_{Physics}(\theta) &= \sum_{x,t \in C} N(u, x, t, u_t, u_x, u_{xx}, \text{...})^2
\end{align}
Here, C represents the set of collocation points on which the Physics-loss is calculated. IC, BC represents the set of points at which Initial Conditions and Boundary Conditions are evaluated. $f$,$h$, represent the initial condition and boundary condition functions respectively. $\theta$ represents the weights of the neural network.

\subsection{Low ranked adaptation for PINNs}
We follow the methodology as proposed in \cite{hu2021lora}. We first train a PINN on task $T_0$ and represent it's weights as $W_0$. Now, we represent the weights of a task $T_1$ as $W_1$ and represent it's update as follows:
\begin{equation}
W_1 = W_0 + \Delta W =  W_0 + A \cdot B\\    
\end{equation}
Here, $W_1\in R^{m\times n}$,$W_0\in R^{m\times n}$, $B\in R^{m\times r}$, $A\in R^{r\times n}$. Here $r<$min\{$m,n$\}. During train time, $W_0$ is kept frozen and matrices $A,B$ contain trainable parameters which adapt from task $T_0$ to task $T_1$.

\subsection{Hypernetworks}
Hypernetworks \cite{ha2016hypernetworks} are neural networks parameterized by $\theta_h$ which accept a task-representation $\lambda$ as input and predict weights $\theta_m$ of a neural network $M$ used to model that particular task.
\begin{align}
\theta_m &= H(\lambda; \theta_h) \nonumber\\  
\textbf{u}(x,t) &= M(x,t;\theta_m)    
\end{align}
Here $\theta_m$ is directly supplied to the main-network for evaluating the solution $\textbf{u}_{pinn}$. We sample different values of $\lambda$ to train the hypernetwork, and generalize to unseen $\lambda$ values at test-time.  

\subsection{LoRA-based Hypernetworks}
LoRA-based Hypernetworks are a combination of Hypernetworks and low-ranked adaptation of the main-network. The mathematical formulation is defined as follows: 
\begin{align}
\theta^\prime_m &= H(\lambda; \theta_h) \nonumber\\
\theta_m &= W_0 + \theta^\prime_m \nonumber\\
\textbf{u}(x,t) &= M(x,t;\theta_m)    
\end{align}
Here, $W_0$ refers to the weights of a pretrained main-network for a task $T_0$. The Hypernetwork $H$, instead of predicting the weights of the entire main-network $M$, just predicts the low ranked weights $A,B$ represented by $\theta^\prime_m = [A:B]$. 

\subsection{HyperPINNs}
HyperPINNs are hypernetworks trained using physics-informed loss introduced in \cite{HyperPINN} to solve parameterized differential equations.
\begin{align}
\theta_m &= H_{pinn} (\lambda; \theta_h) \nonumber\\
\textbf{u}_{pinn}(x,t) &= M_{pinn}(x,t;\theta_m)
\end{align}
HyperPINN consists of two components, a hypernetwork $H_{pinn}$ and a main-network $M_{pinn}$. $H_{pinn}$ takes in task parameterization $\lambda_h$ as input and outputs the weights of the main-network $\theta_m$. 

\subsection{LoRA-based HyperPINNs}

We combine LoRA-based adaptation and HyperPINNs to solve for parameterized PDEs while having to predict a lower number of parameters against a standard HyperPINN. The mathematical formulation of LoRA-based HyperPINN is defined as follows:
\begin{align}
\theta^\prime_m &= H_{pinn}(\lambda; \theta_h) \nonumber\\
\theta_m &= W_0 + \theta^\prime_m \nonumber\\
\textbf{u}(x,t) &= M_{pinn}(x,t;\theta_m)  
\end{align}
We provide an illustration of LoRA-based HyperPINN in figure \ref{fig:HyperPINNLoRA}.
\subsection{Weight regressor hypernetwork}
Instead of using supervised data or physics-informed loss to train the Hypernetwork, Majumdar et al. directly perform regression in weight-space on pretrained PINN weights.
\begin{align}
\theta_m &= H(\lambda; \theta_h) \nonumber\\
\textbf{u}(x,t) &= M(x,t;\theta_m)
\end{align}

Here, the hypernetwork $H$ is trained using $\mathbb{E}_{t \sim T}[(\theta_m - \theta_t)]^2$, where $T$ refers to the task space and $\theta_t$ refers to weights of pretrained PINN for task $t$. 

\subsection{Weight regressor LoRA-based hypernetwork}
Instead of predicting weights of the entire network, Hypernetwork $H$ just predicts the LoRA decomposed weights. Remaining methodology follows from the previous subsection. 
\section{PDE Information}
\subsection{One-dimensional Burger's equation}
The governing PDE system of one-dimensional Burger's equation is given by:
\begin{align}
u_t + u^2_x(x,t)/2 &= \nu u_{xx} \nonumber\\
u(x,0) &= u_0(x)        
\end{align}
Here, $\Omega\in[0,1]$,$\nu=0.01$ refers to the viscosity of the system, and $u_0\in L^2_{per}((0,1); R)$ refers to the initial condition sampled from $\mu$ where $\mu = N(0,625(-\Delta+25I)^{-2})$.
\subsection{Two-dimensional coupled viscous Burger's equation}

The governing equations of two-dimensional coupled viscous Burger's equation is given by:
\begin{align}
u_{t} + u*u_{x} + v*u_{y} &= \nu*(u_{xx} + u_{yy}) \nonumber\\
v_{t} + u*v_{x} + v*v_{y} &= \nu*(v_{xx} + v_{yy})  
\end{align}
The initial and boundary conditions are sampled from the true analytical solutions of the PDE, given by:
\begin{align}
u(x,y,t) &= \frac{3}{4}+\frac{1}{4(1+exp(\frac{(-4x+4y-t)}{32\nu}))} \nonumber\\ v(x,y,t) &= \frac{3}{4}-\frac{1}{4(1+exp(\frac{(-4x+4y-t)}{32\nu}))} 
\end{align}
Here $\Omega\in[0,1]^2$, $t\in[0,1]$, $\nu$ refers to the viscosity of the flow. We consider $\nu\in[1e^{-4},1e^{-3}]$ for our experiments.
\subsection{Navier-Stokes: Kovasznay flow}
Navier-Stokes: Kovasznay flow is a two-dimensional steady state laminar flow. The governing Partial Differential equations are given by:
\begin{align}
u_{x} + v_{y} &= 0\nonumber\\    
u*u_{x} + v*u_{y} &= -p_{x} + (u_{xx} + u_{yy})/Re\nonumber\\
u*v_{x} + v*v_{y} &= -p_{y} + (v_{xx} + v_{yy})/Re    
\end{align}
The boundary conditions for the equation are sampled from the true analytical solutions of the PDE, given by:
\begin{align}
u_{true}&= 1-e^{\lambda x}cos(2\pi y)\nonumber\\
v_{true}&=\lambda e^{\lambda x}sin(2\pi y)/2 \pi\nonumber\\    
p_{true}&=(1-e^{2\lambda x})/2     
\end{align}
Here, $\Omega\in[0,1]^2$, $\lambda=\frac{Re}{2}-\sqrt{\frac{Re^2}{4}+4\pi^2}$, where $Re$ refers to the Reynold's number of the system. We consider $Re\in[20,100]$ for our experiments.

\section{Training and Hyperparameter details}

We take a Multi-layer Perceptron of architecture [inputs-64*4-outputs] for training PINNs. We train a base PINN using an Adam optimizer for 30k epochs with a starting learning rate of 1e-3 for first 10k epochs and use a multiplicative decay of 0.1 with minimum learning rate of $1e^{-7}$ and stop when validation loss doesn't improve for 1000 epochs. We further finetune the trained PINNs using an L-BFGS optimizer. We use the same optimizer and learning-rate schedule for LoRA-based PINNs and finetuning-PINN experiments. For 1D-Burger's equation, we use 128 initial condition points and 10k uniformly sampled collocation points for training the PINNs. In 2D-Burger's equation, we use 400 boundary condition points with 100 points sampled over each face of the square grid, 500 initial condition points and 10k uniformly sampled collocation points. In Navier-Stokes: Kovasznay flow, we consider a $101\times101$ equally spaced grid domain to represent the spatial domain, and sample 2601 collocation
points. We further use 320 boundary condition points, with 80 points
for each face of the grid to enforce the boundary conditions.

In our Hypernetwork-based experiments, we consider an MLP of architecture [inputs- 512*2 - 256*2- 128*2 - outputs]. Here inputs refer to the dimensionality of the task representation, which is 1 for Navier-Stokes (Reynold's number $Re$), 1 for 2D-Burger's (Viscosity $\nu$), and 128 for 1D burger's (Initial condition $u_0(x)$ discretized at 128 positions). Outputs refer to the number of parameters being predicted, with parameter details tabulated in Table \ref{tab:HyperLoRA}. We train all the Hypernetworks using an Adam optimizer for 15k epochs, with a starting learning rate of $1e-3$ for 5k epochs and a multiplicative decay of 0.1 every 3k epochs. Across all examples, we consider 20 validation and 20 test tasks, while we consider 100,20,20 train-tasks for 1D-Burger's, 2D-Burger's and Navier Stokes equations respectively. All experiments were conducted on Nvidia P100 GPU with 16 GB GPU Memory and 1.32 GHz GPU Memory clock using Pytorch framework.
\section{Results and Observations}

\subsection{Rank analysis of decomposed Low-rank matrices}

\begin{table}
    \centering
    \begin{tabular}{|c| c c c c c|}
    \hline
   1D-Burger's &Rank&\#&Error&Time&Epochs\\
\hline
PINNs&&21057&4.54$e^{-6}$&0.28&30k\\
F-PINNs&&21057&1.93$e^{-6}$&0.28&25k\\
\hline
\multirow{7}{*}{L-PINNs}&1&643&1.95$e^{-6}$&0.01&30k\\
&2&1286&1.54$e^{-6}$&0.03&25k\\
&\textbf{4}&\textbf{2572}&\textbf{7.00{$e^{-7}$}}&\textbf{0.05}&\textbf{12k}\\
&8&5144&8.89$e^{-7}$&0.09&12.5k\\
&16&10288&1.00$e^{-6}$&0.12&20k\\
&32&20576&1.03$e^{-6}$&0.16&20k\\
&64&41152&1.15$e^{-6}$&0.28&30k\\
\hline
Kovasznay&Rank&\#&Error&Time&Epochs\\
\hline
PINNs&&21187&1.05$e^{-6}$&0.22&30k\\
F-PINNs&&21187&6.36$e^{-7}$&0.22&20k\\
\hline
\multirow{7}{*}{L-PINNs}&1&645&8.16$e^{-7}$&0.02&30k\\
&2&1290&6.68$e^{-7}$&0.04&20k\\
&\textbf{4}&\textbf{2580}&\textbf{3.85$e^{-7}$}&\textbf{0.06}&\textbf{12k}\\
&8&5160&5.59$e^{-7}$&0.09&12.5k\\
&16&10320&9.35$e^{-7}$&0.14&20k\\
&32&20640&9.74$e^{-7}$&0.16&30k\\
&64&41280&1.18$e^{-6}$&0.22&30k\\
\hline
2D-Burger's&Rank&\#&Error&Time&Epochs\\
\hline
PINNs&&21187&3.25$e^{-5}$&0.33&30k\\
F-PINNs&&21187&2.69$e^{-5}$&0.33&25.5k\\
\hline
\multirow{7}{*}{L-PINNs}&1&645&3.81$e^{-4}$&0.02&30k\\
&2&1290&7.02$e^{-5}$&0.04&25k\\
&\textbf{4}&\textbf{2580}&\textbf{1.17$e^{-5}$}&\textbf{0.05}&\textbf{12k}\\
&8&5160&2.73$e^{-5}$&0.10&13k\\
&16&10320&2.89$e^{-5}$&0.14&20k\\
&32&20640&2.94$e^{-5}$&0.17&20k\\
&64&41280&3.01$e^{-5}$&0.36&30k\\
\hline
    \end{tabular}
    \caption{Low-ranked Adaptation results. Here, F-PINNs refer to PINNs finetuned directly, L-PINNs refer to Low-ranked adaptation based PINNs. Error refers to MSE averaged over 20 test-tasks, to be used in Table \ref{tab:HyperLoRA}. Time refers to training time per epoch in seconds. \# refers to number of parameters updated in training-time.}
    \label{tab:LoRA}
\end{table}

In this subsection, we analyze how the performance varies by changing the rank of our decomposed matrices of Neural network used for low-ranked adaptation. We tabulate the results of our Low-ranked adaptation experiments in Table \ref{tab:LoRA}. Our experimental design is as follows: Across all PDE examples, we first train an independent PINN on a task $T_0$. Then we perform Low-ranked adaptation for a new task $T_x$ using the trained PINN for task $T_0$ as our base model. Lower the rank of the decomposed tensors, the fewer the number of parameters involved in adapting to the newer task. 

Across all three PDE systems, we notice the average accuracy across the test-tasks first increases with increase in rank, reaches an optimal rank, which happens to be 4, and then slightly deteriorates as the rank is further increased to the full rank of the original matrix. The best LoRA-based PINN is on average 3.98$\times$ better than PINNs, while requiring $8.22\times$ fewer parameters, $2.5\times$ fewer epochs while  yielding $5.25\times$ faster training time per epoch (due to training fewer parameters). Additionally, the best LoRA-based PINN is on average $2.23\times$ better than a finetuned-PINN, while requiring $8.22\times$ fewer parameters, $2\times$ fewer epochs yielding $5.25\times$ faster training time per epoch. 

Improvement of performance due to initial rank increase can be attributed to improved representation capacity of the finetuned task $T_x$ due to higher number of parameters. We see a 2.78\% improvement in 1D-Burger's, 2.11\% improvement in Navier-Stokes, and 32.56\% improvement in 2D-Burger's on increasing the rank from 1 to 4. We additionally notice, while the training-time per epoch is lower for LoRA-based PINNs with ranks 1 and 2, as the representation capacity is restricted, these take a higher time to converge as compared to LoRA-based PINNs with rank 4. Further increase in rank leads to drop in performance, and we speculate overfitting and increased in variance to be reasons for it due to increase in \# parameters. In addition, very high ranked LoRA-based PINNs require higher training time to converge, as against the best-ranked LoRA-based PINN.

\subsection{Hypernetworks for Low-ranked adaptation}

\begin{figure*}
    \centering
\includegraphics[height=5cm, width=18cm]{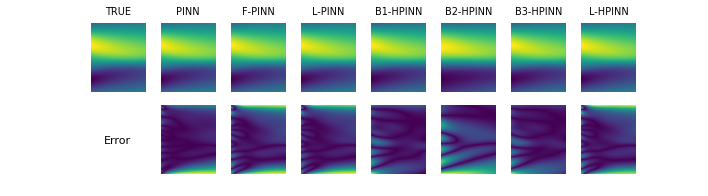}
    \caption{Predictions and error maps of Burger's equation. F-PINN refers to finetuned PINNs, L-PINN refers to LoRA-based PINNs, B1,B2,B3 refers to benchmarks 1,2,3 of Hypernetworks, and L-HPINN refers to LoRA-based HyperPINN.}
    \label{fig:Burgers}
\end{figure*}

\begin{figure*}
    \centering
\includegraphics[height=6cm, width=18cm]{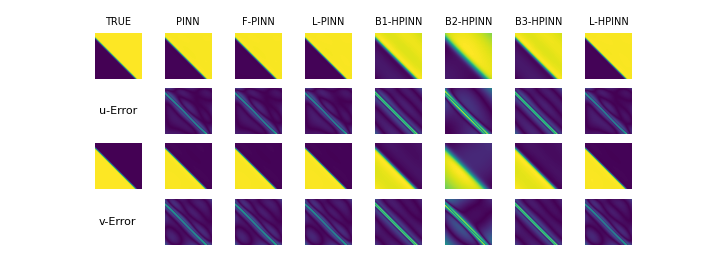}
    \caption{Predictions and error maps of 2D Burger's equation. Terminologies follow from Figure \ref{fig:Burgers}.}
    \label{fig:2DBurgers}
\end{figure*}

\begin{figure*}
    \centering
\includegraphics[height=8cm, width=18cm]{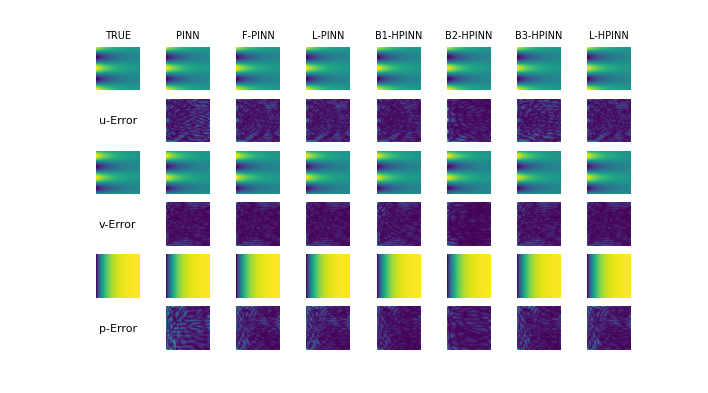}
    \caption{Predictions and error maps of Navier-Stokes: Kovasznay flow. Terminologies follow from Figure \ref{fig:Burgers}.}
    \label{fig:Navier-Stokes}
\end{figure*}

\begin{table*}
\centering
    \begin{tabular}{|c| c| c c c c c c|}
    \hline
    	&&Rank& \# Parameters&Train&Valid&Test&Time per epoch\\
     \cline{3-8}
&&1&645&2.19$e^{-6}$&1.39$e^{-6}$&1.87$e^{-6}$&0.16\\
&&2&1290&4.45$e^{-7}$&9.16$e^{-7}$&8.31$e^{-7}$&0.21\\
&Hypernetwork+LoRA&\textbf{4}&\textbf{2580}&\textbf{4.08$e^{-7}$}&\textbf{5.16$e^{-7}$}&\textbf{5.23$e^{-7}$}&\textbf{0.29}\\
&&8&5160&6.46$e^{-7}$&6.99$e^{-7}$&7.75$e^{-7}$&0.37\\
&&*&21187&4.38$e^{-4}$&3.60$e^{-3}$&2.66$e^{-3}$&2.15\\
\cline{2-8}
&&1&645&4.17$e^{-3}$&3.29$e^{-3}$&3.22$e^{-3}$&0.05\\
&&2&1290&3.95$e^{-3}$&3.59$e^{-3}$&3.49$e^{-3}$&0.07\\
&Hypernetwork+LoRA&\textbf{4}&\textbf{2580}&\textbf{2.95$e^{-3}$}&\textbf{3.06$e^{-3}$}&\textbf{3.19$e^{-3}$}&\textbf{0.101}\\
&(Weight Regression)&8&5160&4.04$e^{-3}$&5.19$e^{-3}$&7.72$e^{-3}$&0.12\\
Navier-Stokes&&*&21187&1.17$e^{-2}$&2.18$e^{-2}$&1.95$e^{-2}$&0.73\\
\cline{2-8}
Kovasznay flow&&1&645&2.19$e^{-6}$&1.39$e^{-6}$&1.91$e^{-6}$&0.16\\
&&2&1290&4.28$e^{-7}$&9.03$e^{-7}$&8.29$e^{-7}$&0.21\\
&Hypernetwork+LoRA&\textbf{4}&\textbf{2580}&\textbf{4.11$e^{-7}$}&\textbf{5.04$e^{-7}$}&\textbf{5.26$e^{-7}$}&\textbf{0.29}\\
&(PINN generated)&8&5160&6.28$e^{-7}$&7.15$e^{-7}$&7.64$e^{-7}$&0.37\\
&&*&21187&4.47$e^{-4}$&3.66$e^{-3}$&2.82$e^{-3}$&2.15\\
\cline{2-8}
&&1&645&2.22$e^{-6}$&1.35$e^{-6}$&1.80$e^{-6}$&0.63\\
&&2&1290&4.36$e^{-7}$&9.15$e^{-7}$&8.25$e^{-7}$&0.84\\
&HyperPINN+LoRA&4&\textbf{2580}&\textbf{4.15$e^{-7}$}&\textbf{5.52$e^{-7}$}&\textbf{4.84$e^{-7}$}&\textbf{1.15}\\
&&8&5160&5.93$e^{-7}$&6.64$e^{-7}$&7.22$e^{-7}$&1.45\\
&&*&21187&3.17$e^{-4}$&2.91$e^{-3}$&4.42$e^{-3}$&9.43\\								
 \hline
\multirow{20}{*}{1D Burger's} &&1&643&7.25$e^{-4}$&1.09$e^{-3}$&8.78$e^{-4}$&1.19\\
&&2&1286&2.69$e^{-4}$&4.67$e^{-4}$&4.98$e^{-4}$&1.56\\
&Hypernetwork+LoRA&\textbf{4}&\textbf{2572}&\textbf{3.02$e^{-4}$}&\textbf{3.47$e^{-4}$}&\textbf{3.22$e^{-4}$}&\textbf{1.84}\\
&&8&5144&5.69$e^{-4}$&6.11$e^{-4}$&6.37$e^{-4}$&2.25\\
&&*&21057&1.96$e^{-2}$&3.70$e^{-2}$&3.14$e^{-2}$&17.26\\
\cline{2-8}
&&1&643&8.82$e^{-3}$&8.54$e^{-3}$&9.13$e^{-3}$&0.81\\
&&2&1286&6.19$e^{-3}$&6.15$e^{-3}$&6.15$e^{-3}$&1.05\\
&Hypernetwork+LoRA&\textbf{4}&\textbf{2572}&\textbf{4.24$e^{-3}$}&\textbf{4.43$e^{-3}$}&\textbf{4.17$e^{-3}$}&\textbf{1.36}\\
&(Weight regression)&8&5144&5.43$e^{-3}$&5.94$e^{-3}$&5.56$e^{-3}$&1.74\\
&&*&21057&3.64$e^{-2}$&4.90$e^{-2}$&5.12$e^{-2}$&10.14\\
\cline{2-8}
&&1&643&7.18$e^{-4}$&1.12$e^{-3}$&8.89$e^{-4}$&1.19\\
&&2&1286&2.86$e^{-4}$&4.75$e^{-4}$&5.05$e^{-4}$&1.56\\
&Hypernetwork+LoRA&\textbf{4}&\textbf{2572}&\textbf{3.12$e^{-4}$}&\textbf{3.55$e^{-4}$}&\textbf{3.35$e^{-4}$}&\textbf{1.84}\\
&(PINN generated)&8&5144&5.58$e^{-4}$&6.16$e^{-4}$&6.44$e^{-4}$&2.25\\
&&*&21057&2.02$e^{-2}$&3.84$e^{-2}$&3.26$e^{-2}$&17.26\\
\cline{2-8}
&&1&643&9.64$e^{-6}$&1.45$e^{-5}$&1.17$e^{-5}$&3.15\\
&&2&1286&4.02$e^{-6}$&6.22$e^{-6}$&6.64$e^{-6}$&4.35\\
&HyperPINN+LoRA&\textbf{4}&\textbf{2572}&\textbf{3.67$e^{-6}$}&\textbf{4.82$e^{-6}$}&\textbf{4.74$e^{-6}$}&\textbf{5.75}\\
&&8&5144&7.19$e^{-6}$&8.19$e^{-6}$&8.23$e^{-6}$&8.45\\
&&*&21057&2.58$e^{-3}$&4.29$e^{-3}$&3.97$e^{-3}$&67.6\\						
 \hline
\multirow{20}{*}{2D Burger's}&&1&644&3.05$e^{-4}$&4.44$e^{-4}$&9.52$e^{-4}$&0.093\\
&&2&1288&2.97$e^{-4}$&3.26$e^{-4}$&2.55$e^{-4}$&0.131\\
&Hypernetwork+LoRA&\textbf{4}&\textbf{2576}&\textbf{1.46$e^{-4}$}&\textbf{1.50$e^{-4}$}&\textbf{1.16$e^{-4}$}&\textbf{0.171}\\
&&8&5152&1.81$e^{-4}$&1.66$e^{-4}$&1.52$e^{-4}$&0.241\\
&&*&21122&1.34$e^{-2}$&2.15$e^{-2}$&3.97$e^{-2}$&1.91\\
\cline{2-8}
&&1&644&3.17$e^{-2}$&3.22$e^{-2}$&2.68$e^{-2}$&0.031\\
&&2&1288&2.12$e^{-2}$&2.15$e^{-2}$&2.36$e^{-2}$&0.044\\
&Hypernetwork+LoRA&\textbf{4}&\textbf{2576}&\textbf{1.40$e^{-2}$}&\textbf{1.57$e^{-2}$}&\textbf{2.15$e^{-2}$}&\textbf{0.059}\\
&(Weight Regression)&8&5152&2.66$e^{-2}$&2.36$e^{-2}$&2.94$e^{-2}$&0.082\\
&&*&21122&1.06$e^{-2}$&3.32$e^{-2}$&1.00$e^{-2}$&2.95\\
\cline{2-8}
&&1&644&3.12$e^{-4}$&4.63$e^{-4}$&9.58$e^{-4}$&0.093\\
&&2&1288&2.88$e^{-4}$&3.32$e^{-4}$&2.57$e^{-4}$&0.131\\
&Hypernetwork+LoRA &\textbf{4}&\textbf{2576}&\textbf{1.49$e^{-4}$}&\textbf{1.61$e^{-4}$}&\textbf{1.19$e^{-4}$}&\textbf{0.171}\\
&(PINN generated)&8&5152&1.85$e^{-4}$&1.58$e^{-4}$&1.48$e^{-4}$&0.241\\
&&*&21122&1.34$e^{-2}$&2.14$e^{-2}$&3.88$e^{-2}$&1.91\\
\cline{2-8}
&&1&644&5.52$e^{-4}$&8.26$e^{-4}$&7.35$e^{-4}$&0.375\\
&&2&1288&6.14$e^{-5}$&7.66$e^{-5}$&8.22$e^{-5}$&0.525\\
&HyperPINN+LoRA&\textbf{4}&\textbf{2576}&\textbf{2.89$e^{-5}$}&\textbf{4.26$e^{-5}$}&\textbf{4.19$e^{-5}$}&\textbf{0.69}\\
&&8&5152&2.72$e^{-5}$&4.31$e^{-5}$&4.18$e^{-5}$&0.975\\
&&*&21122&4.52$e^{-3}$&1.91$e^{-2}$&2.05$e^{-2}$&7.85\\
\hline
    \end{tabular}
    \caption{Mean squared error of Hypernetwork-based LoRA experiments}
    \label{tab:HyperLoRA}
\end{table*}
							
In this subsection, we study the results for our experiments for LoRA-based HyperPINNs. Hypernetwork-based PINN architectures have faster inference time than PINN-based counterparts, and allows for generalization to parameterized PDE systems. We tabulate the results in Table \ref{tab:HyperLoRA}. Across all examples, * refers to the case where we don't perform LoRA, and hypernetwork predicts the entire neural network. 

As our first benchmark, we train a LoRA-based hypernetwork using the true simulation data and solutions of the PDE as ground-truth. Similar to our analysis in previous subsection, we observe, across all PDE examples, the Hypernetwork performance improves till LoRA of rank 4, and then starts deteriorating. The reason for improved performance for rank 4 over lower ranks is higher representation capacity of base-networks. The reason for drop in performance for Hypernetworks beyond rank 4 is inability of Hypernetworks to predict very high dimensional weights in a regression setting. However, on test tasks, Rank-4 LoRA-based hypernetworks are just $1.35\times$ worse than Rank-4 LoRA-based PINNs on Kovasznay flow, but are $10.15\times$ and $460\times$ worse on 2D-Burger's and 1D-Burger's respectively. Thus, this benchmark has two issues. First, this benchmark requires pre-existing simulation data for multiple solved PDE instances which may not be available in practice, and second, the generalization is poor and physics is violated as seen in Figures \ref{fig:Burgers}, \ref{fig:2DBurgers}, \ref{fig:Navier-Stokes}. 

We mitigate the data availability drawback by training multiple instances of PINNs and use the weights of the data generated by trained-PINNs as supervised labels for benchmark 2. In our second benchmark, we use the hypernetwork to predict the weights of the base-network and use weights of the pre-trained PINNs as regression loss following motivations from the paper \cite{majumdar2022realtime}. While the rank analysis remains similar and the hypernetwork has lower training time, the performace across all test-examples is quite poor. Reason: Weight-spaces are extremely sensitive to slight perturbations, thus leading to very complex manifolds being learnt. The complexity of generalizing the output manifold to test-examples is increased, thus leading to poor performance. Weight-regressor hypernetworks typically perform 2-3 orders worse than LoRA-based PINNs and benchmark 1.

In benchmark 3, we use the data generated by trained-PINNs as supervised labels and train a hypernetwork. We observe, the rank analysis to be similar as in the past sections, and the best-ranked hypernetwork performs  comparably ($1.02\times$ worse) to the best-ranked hypernetwork in benchmark 1, and is 3-4 orders superior in performance as compared to weight-regressor hypernetworks. However, this benchmark like benchmark 1 performs similarly worse  ($1.36\times$,$4.19\times$ and $463\times$)  than LoRA-based PINNs of rank 4 on Kovasznay flow, 2D Burger's and 1D Burger's respectively. As a summary, this benchmark has 2 issues: 1) Training multiple PINNs can be costly  2) Poor generalization and violation of physics constraints, as seen in Figure \ref{fig:Burgers}, \ref{fig:2DBurgers}, \ref{fig:Navier-Stokes}. 

We mitigate both the issues in benchmark 3 by using LoRA-based HyperPINNs. HyperPINNs take in a task embedding as input, and learn the hypernetworks in a physics-informed manner, instead of taking pre-trained data as a supervision signal. This eradicates the need of having ground-truth samples or neural surrogate samples like in benchmark 3. We observe the performance of Rank 4 LoRA-based HyperPINN is 3 and 2 orders of magnitude better in 1D-Burger's equation and 2D-Burger's equation respectively, and is less-than an order ($3.61\times$) worse than LoRA-based PINNs on test-examples. This indicates LoRA-based HyperPINNs have the capability to generalize across test-tasks at an inference-time advantage, compared to traditional PINNs and LoRA-based PINNs, while respecting the physics constraints observed in Figures \ref{fig:Burgers}, \ref{fig:2DBurgers}, \ref{fig:Navier-Stokes}.

We tabulate the inference time for our architectures in Table \ref{tab:inference-time}. Across all examples, we observe LoRA-based HyperPINNs to be around 2-3 orders of magnitude faster than vanilla architectures, as the entire computational cost is transferred to one-time train cost, and we simply predict the weights for a test-time, without having to finetune. The best-ranked LoRA-based PINNs are $10.25\times$ faster than finetuned PINNs and $13.21\times$ faster than vanilla-PINNs, due to training lower number of parameters and incorporating pre-trained information from an already trained base-task.
\begin{table}
    \centering
    \begin{tabular}{c c c}
        &Architecture & Time  \\
        &&(sec)\\
        \hline
        \multirow{4}{*}{1D-Burger's}&PINN & 8400 \\
        &Finetuned PINN & 7000\\
        &LoRA-based PINN & 600\\
        &LoRA-based HyperPINN & 1.1\\
        \hline
        &PINN & 6600 \\
        Navier-Stokes&Finetuned PINN & 4400\\
        (Kovasznay flow)&LoRA-based PINN & 720\\
        &LoRA-based HyperPINN & 1.2\\
        \hline
        \multirow{4}{*}{2D-Burger's}&PINN & 9900 \\
        &Finetuned PINN & 8415\\
        &LoRA-based PINN & 602\\
        &LoRA-based HyperPINN & 1.2\\
        \hline
    \end{tabular}
    \caption{Inference-time}
    \label{tab:inference-time}
\end{table}

\section{Conclusions}

We use Low-ranked adaptation for PINNs to quickly adapt solutions of parameterized PDEs from one instance to another. We investigate the importance of choice of rank of decomposed tensors, and conclude there exists an optimal rank for tensor decomposition, lowering which leads to reduced representation capacity and larger training time for the newer PDE instance. Increasing the rank leads to overfitting and a slight drop in performance at test-time. We note, the optimal low-rank adapted PINN converges faster and outperforms PINNs trained from random initialization or finetuned PINNs. Next, we scale LoRA-based PINNs to HyperLoRA-based PINNs to further reduce inference time. Rank analysis of LoRA-based PINNs extend to LoRA-based Hypernetworks as well. Benchmark LoRA-based hypernetworks have the drawback of requiring pre-existing ground-truth data which is scarce, and have inferior generalization as the underlying physics isn't captured. Utilizing a physics-loss to train the LoRA-based Hypernetworks leads to improved generalization and comparable performance with instance-wise LoRA-based PINNs, while retaining the advantage in inference-time. 

\section{Limitations and future work}
One of the limitations of LoRA-based PINNs, is the manual search involved in determining the best rank of the matrix decomposition. Efficiently determining the best rank alongside theoritical guarantees will considerably help reduce manual experimentation. Second drawback, PINNs fail to model long temporal flows simply using an unsupervised physics loss, and thus we speculate LoRA-based HyperPINNs will also have difficulties modeling long temporal flows. We seek to analyze how LoRA-based HyperPINNs will scale to these complex examples where the complex loss-landscape of physics-loss will play a big factor during convergence at training time.  

\bibliography{aaai24}

\begin{thebibliography}{16}
\providecommand{\natexlab}[1]{#1}

\bibitem[{de~Avila Belbute-Peres, fan Chen, and Sha(2021)}]{HyperPINN}
de~Avila Belbute-Peres, F.; fan Chen, Y.; and Sha, F. 2021.
\newblock HyperPINN: Learning parameterized differential equations with
  physics-informed hypernetworks.
\newblock arXiv:2111.01008.

\bibitem[{Dupont et~al.(2022)Dupont, Kim, Eslami, Rezende, and
  Rosenbaum}]{dupont2022data}
Dupont, E.; Kim, H.; Eslami, S. M.~A.; Rezende, D.; and Rosenbaum, D. 2022.
\newblock From data to functa: Your data point is a function and you can treat
  it like one.
\newblock arXiv:2201.12204.

\bibitem[{Greenfeld et~al.(2019)Greenfeld, Galun, Basri, Yavneh, and
  Kimmel}]{greenfeld2019learning}
Greenfeld, D.; Galun, M.; Basri, R.; Yavneh, I.; and Kimmel, R. 2019.
\newblock Learning to optimize multigrid PDE solvers.
\newblock In \emph{International Conference on Machine Learning}, 2415--2423.
  PMLR.

\bibitem[{Guo, Li, and Iorio(2016)}]{10.1145/2939672.2939738}
Guo, X.; Li, W.; and Iorio, F. 2016.
\newblock Convolutional Neural Networks for Steady Flow Approximation.
\newblock In \emph{Proceedings of the 22nd ACM SIGKDD International Conference
  on Knowledge Discovery and Data Mining}, KDD '16, 481–490. New York, NY,
  USA: Association for Computing Machinery.
\newblock ISBN 9781450342322.

\bibitem[{Ha(2016)}]{ha2016generating}
Ha, D. 2016.
\newblock Generating Large Images from Latent Vectors.
\newblock \emph{blog.otoro.net}.

\bibitem[{Ha, Dai, and Le(2016)}]{ha2016hypernetworks}
Ha, D.; Dai, A.; and Le, Q.~V. 2016.
\newblock HyperNetworks.
\newblock arXiv:1609.09106.

\bibitem[{Hu et~al.(2021)Hu, Shen, Wallis, Allen-Zhu, Li, Wang, Wang, and
  Chen}]{hu2021lora}
Hu, E.~J.; Shen, Y.; Wallis, P.; Allen-Zhu, Z.; Li, Y.; Wang, S.; Wang, L.; and
  Chen, W. 2021.
\newblock LoRA: Low-Rank Adaptation of Large Language Models.
\newblock arXiv:2106.09685.

\bibitem[{Ivison et~al.(2023)Ivison, Bhagia, Wang, Hajishirzi, and
  Peters}]{ivison2023hint}
Ivison, H.; Bhagia, A.; Wang, Y.; Hajishirzi, H.; and Peters, M. 2023.
\newblock HINT: Hypernetwork Instruction Tuning for Efficient Zero- Few-Shot
  Generalisation.
\newblock arXiv:2212.10315.

\bibitem[{Kochkov et~al.(2021)Kochkov, Smith, Alieva, Wang, Brenner, and
  Hoyer}]{kochkov2021machine}
Kochkov, D.; Smith, J.~A.; Alieva, A.; Wang, Q.; Brenner, M.~P.; and Hoyer, S.
  2021.
\newblock Machine learning--accelerated computational fluid dynamics.
\newblock \emph{Proceedings of the National Academy of Sciences}, 118(21):
  e2101784118.

\bibitem[{Li and Liang(2021)}]{li2021prefixtuning}
Li, X.~L.; and Liang, P. 2021.
\newblock Prefix-Tuning: Optimizing Continuous Prompts for Generation.
\newblock arXiv:2101.00190.

\bibitem[{Li et~al.(2020)Li, Kovachki, Azizzadenesheli, Liu, Bhattacharya,
  Stuart, and Anandkumar}]{li2020fourier}
Li, Z.; Kovachki, N.; Azizzadenesheli, K.; Liu, B.; Bhattacharya, K.; Stuart,
  A.; and Anandkumar, A. 2020.
\newblock Fourier neural operator for parametric partial differential
  equations.
\newblock \emph{arXiv preprint arXiv:2010.08895}.

\bibitem[{Lu, Jin, and Karniadakis(2019)}]{lu2019deeponet}
Lu, L.; Jin, P.; and Karniadakis, G.~E. 2019.
\newblock Deeponet: Learning nonlinear operators for identifying differential
  equations based on the universal approximation theorem of operators.
\newblock \emph{arXiv preprint arXiv:1910.03193}.

\bibitem[{Majumdar et~al.(2022)Majumdar, Jadhav, Deodhar, Karande, Vig, and
  Runkana}]{majumdar2022realtime}
Majumdar, R.; Jadhav, V.; Deodhar, A.; Karande, S.; Vig, L.; and Runkana, V.
  2022.
\newblock Real-time Health Monitoring of Heat Exchangers using Hypernetworks
  and PINNs.
\newblock arXiv:2212.10032.

\bibitem[{Mescheder et~al.(2019)Mescheder, Oechsle, Niemeyer, Nowozin, and
  Geiger}]{Mescheder_2019_CVPR}
Mescheder, L.; Oechsle, M.; Niemeyer, M.; Nowozin, S.; and Geiger, A. 2019.
\newblock Occupancy Networks: Learning 3D Reconstruction in Function Space.
\newblock In \emph{Proceedings of the IEEE/CVF Conference on Computer Vision
  and Pattern Recognition (CVPR)}.

\bibitem[{Raissi, Perdikaris, and Karniadakis(2019)}]{RAISSI2019686}
Raissi, M.; Perdikaris, P.; and Karniadakis, G. 2019.
\newblock Physics-informed neural networks: A deep learning framework for
  solving forward and inverse problems involving nonlinear partial differential
  equations.
\newblock \emph{Journal of Computational Physics}, 378: 686--707.

\bibitem[{Ruiz et~al.(2023)Ruiz, Li, Jampani, Wei, Hou, Pritch, Wadhwa,
  Rubinstein, and Aberman}]{ruiz2023hyperdreambooth}
Ruiz, N.; Li, Y.; Jampani, V.; Wei, W.; Hou, T.; Pritch, Y.; Wadhwa, N.;
  Rubinstein, M.; and Aberman, K. 2023.
\newblock HyperDreamBooth: HyperNetworks for Fast Personalization of
  Text-to-Image Models.
\newblock \emph{arXiv preprint arXiv:2307.06949}.

\end{thebibliography}

\end{document}